\documentclass[letterpaper, 10 pt, conference]{ieeeconf}  

\IEEEoverridecommandlockouts                              

\usepackage{amsmath} 
\usepackage{amssymb}  
\usepackage{graphicx}
\usepackage[export]{adjustbox}
\usepackage{array}
\usepackage{multirow}
\usepackage{tabularx}
\usepackage{support-caption}
\usepackage{subcaption}
\usepackage[utf8]{inputenc}
\usepackage[export]{adjustbox}
\usepackage{wrapfig}
\usepackage{threeparttable}
\usepackage[english]{babel}
\usepackage{color, colortbl}
\usepackage{makecell}

\title{\LARGE \bf
CLOCs: Camera-LiDAR Object Candidates Fusion\\ for 3D Object Detection 
}

\author{Su Pang, Daniel Morris, Hayder Radha
\thanks{Su Pang, Daniel Morris and Hayder Radha are with the Department of Electrical and Computer Engineering, College of Engineering, Michigan State University, 220 Trowbridge Road, East Lansing, Michigan, 48824, United States. Email: {\tt\small pangsu@msu.edu, dmorris@msu.edu, radha@egr.msu.edu}}%
}

\begin{document}
\maketitle
\thispagestyle{empty}
\pagestyle{empty}

\begin{abstract}

There have been significant advances in neural networks for both 3D object detection using LiDAR and 2D object detection using video.  However, it has been surprisingly difficult to train networks to effectively use both modalities in a way that demonstrates gain over single-modality networks.  In this paper, we propose a novel Camera-LiDAR Object Candidates (CLOCs) fusion network. CLOCs fusion  provides a low-complexity multi-modal fusion framework that  significantly improves the performance of single-modality detectors. CLOCs operates on the combined output candidates before Non-Maximum Suppression (NMS) of any 2D and any 3D detector, and is trained to leverage their geometric and semantic consistencies to produce more accurate final 3D and 2D detection results. Our experimental evaluation on the challenging KITTI object detection benchmark, including 3D and bird's eye view metrics, shows significant improvements, especially at long distance, over the state-of-the-art fusion based methods. At time of submission, CLOCs ranks the highest among all the fusion-based methods in the official KITTI leaderboard. We will release our code upon acceptance.

\end{abstract}

\section{INTRODUCTION}

\begin{figure}[t]
\centering
    \begin{subfigure}{0.49\columnwidth}
        \includegraphics[width=\columnwidth,center]{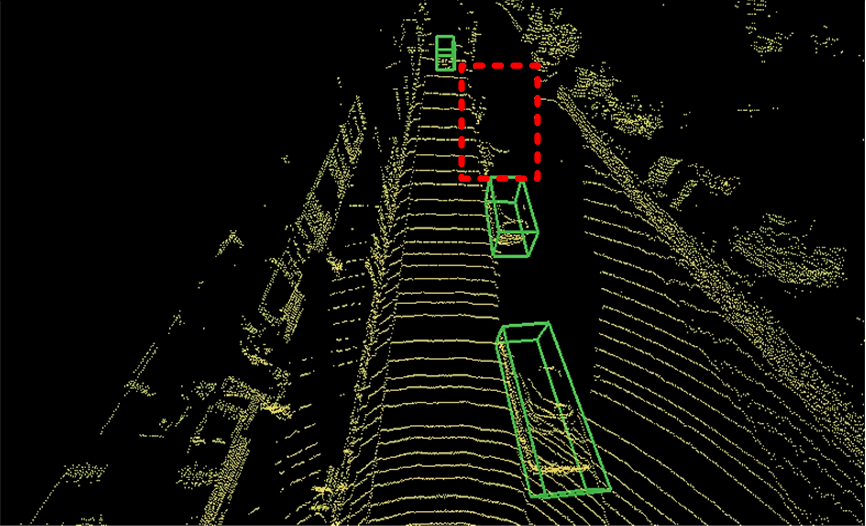}
        \caption{LiDAR-only detector on \#1}
        \label{figure 1a}
    \end{subfigure}
    \begin{subfigure}{0.49\columnwidth}
        \includegraphics[width=\columnwidth,center]{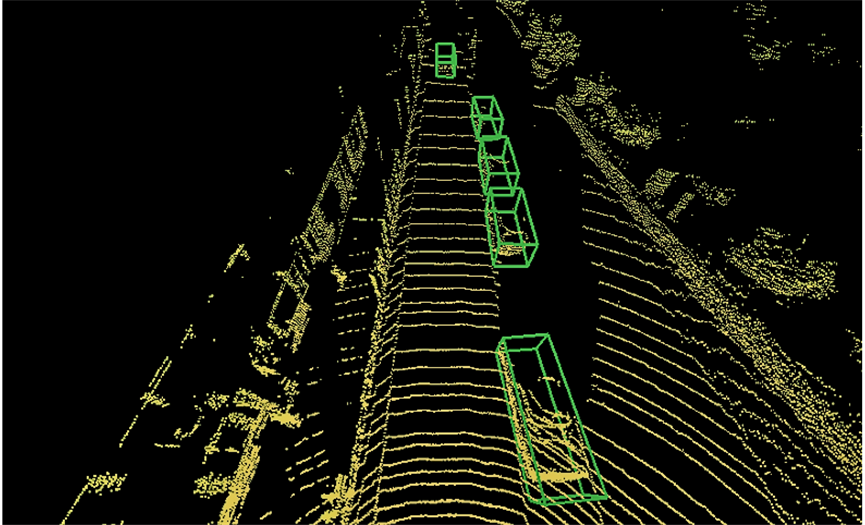}
        \caption{Our CLOCs fusion on \#1}
        \label{figure 1b}
    \end{subfigure}
    
    \begin{subfigure}{0.49\columnwidth}
        \includegraphics[width=\columnwidth,center]{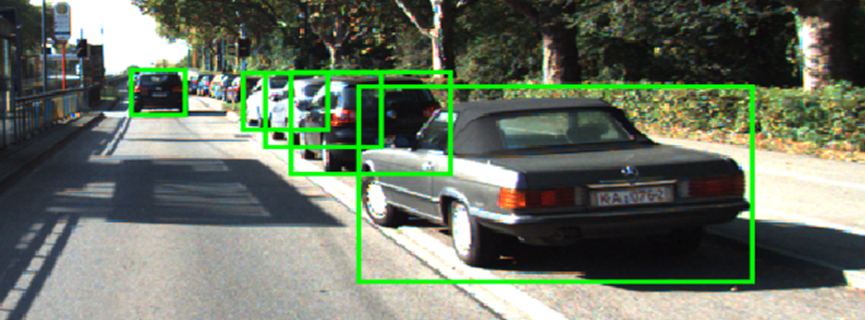}
        \caption{Image-only detector on \#1}
        \label{figure 1c}
    \end{subfigure}
    \begin{subfigure}{0.49\columnwidth}
        \includegraphics[width=\columnwidth,center]{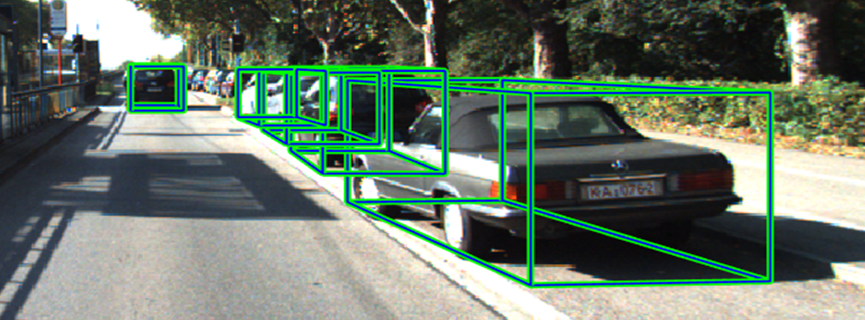}
        \caption{CLOCs on \#1 shown in img}
        \label{figure 1d}
    \end{subfigure}
    
    \begin{subfigure}{0.99\columnwidth}
        \includegraphics[width=\columnwidth,height=2cm]{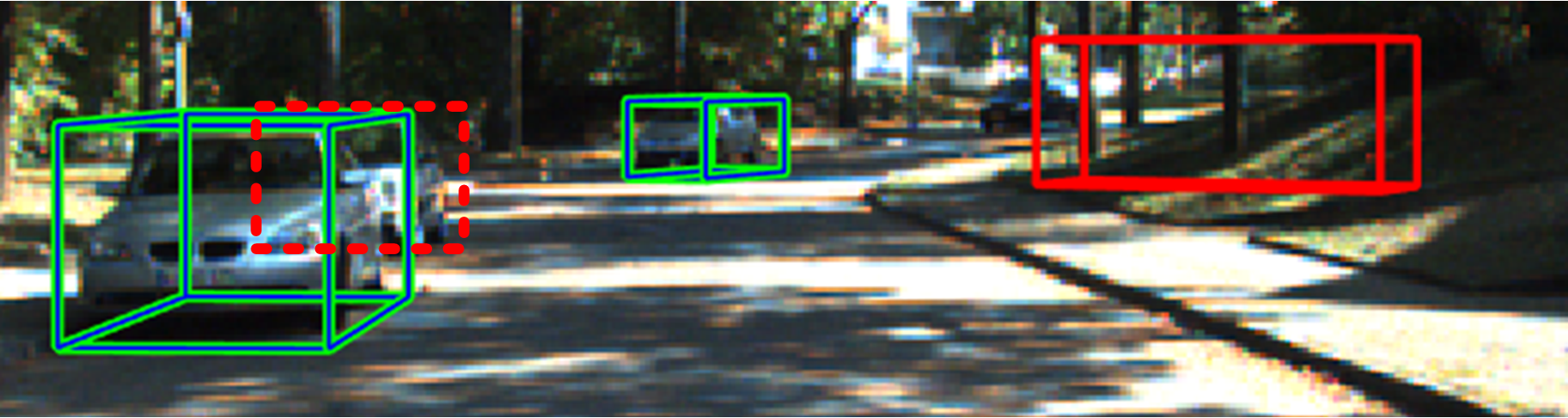}
        \caption{LiDAR-only detector on \#2 shown in image}
        \label{figure 1e}
    \end{subfigure}
    
    \begin{subfigure}{0.99\columnwidth}
        \includegraphics[trim=5 0 0 0,clip,width=\columnwidth,height=2cm]{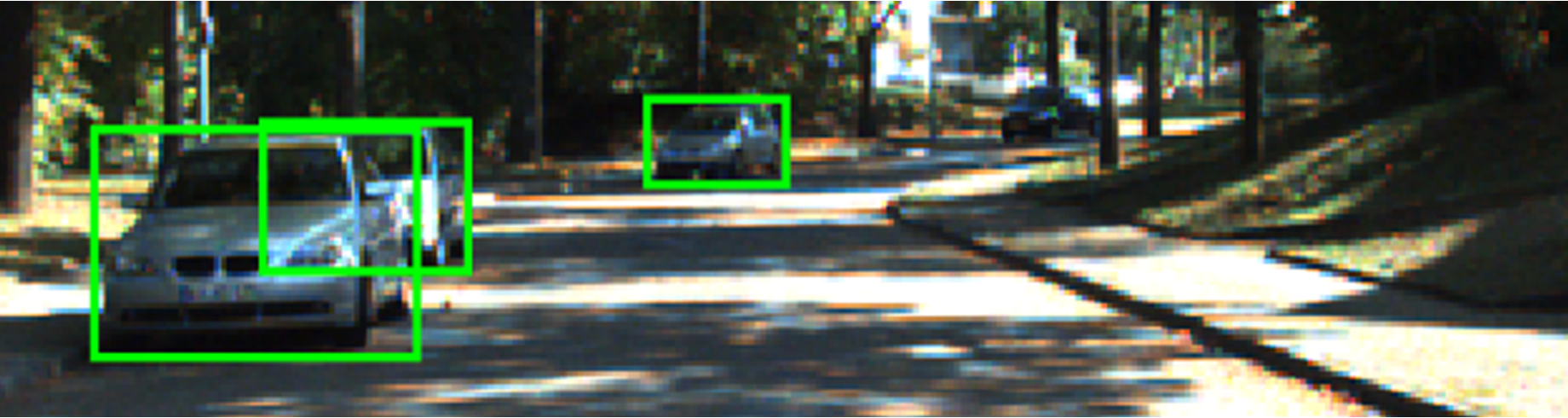}
        \caption{Image-only detector on \#2}
        \label{figure 1f}
    \end{subfigure}
    
    \begin{subfigure}{0.99\columnwidth}
        \includegraphics[width=\columnwidth,height=2cm]{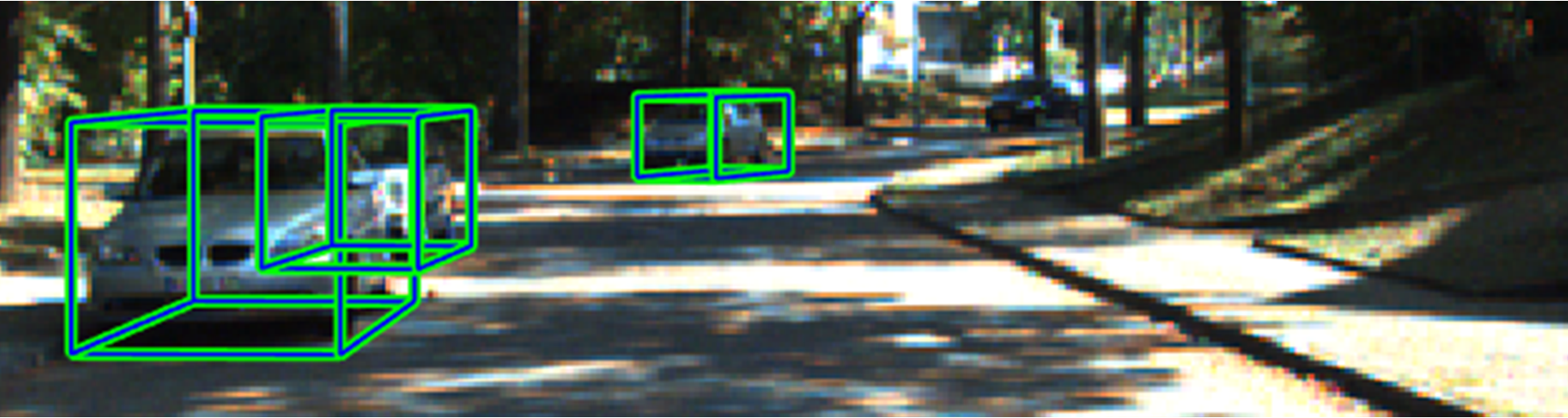}
        \caption{Our CLOCs fusion on \#2 shown in image}
        \label{figure 1g}
    \end{subfigure}
    
\caption{Example \#1 of car detection from single modality methods: (a) LiDAR-only detector, and (c) image-only detector, with our CLOCs fusion shown in (b) and (d). A second example, \#2, is shown below. Dashed red box shows misses and solid red bounding box shows false positives.  Our proposed CLOCs fusion can correct both of these errors.}
\vspace{-0.5cm}
\label{figure1}
\end{figure}

Autonomous driving systems need accurate 3D perception of vehicles and other objects in their environment. Unlike  2D visual detection, 3D-based object detection enables spatial path planning for object avoidance and navigation.  Compared to 2D object detection, which has been well-studied ~\cite{ren2015faster,Calandra2016,lin2017feature,lin2017focal}, 3D object detection is more challenging with more output parameters needed to specify 3D oriented bounding boxes around targets.  In addition, LiDAR methods~\cite{zhou2018voxelnet,yan2018second,qi2017pointnet,shi2019pointrcnn,lang2019pointpillars} are hampered by typically lower input data resolution than video which has a large adverse impact on accuracy at longer ranges. Fig.~\ref{figure1} illustrates the difficulty in detecting vehicles from just a few points and no texture at long range.  Human annotators use both the camera images together with the LiDAR point clouds to create the ground truth bounding boxes~\cite{geiger2012we}. This motivates multi-modal sensor fusion as a way to improve single-modal methods.

While sensor fusion has potential to address the shortcomings of video-only and LiDAR-only detections, finding an effective approach that improves on the state-of-the-art single modality detectors has been difficult.  This is illustrated in the official KITTI 3D object detection benchmark leaderboard, where LiDAR-only based methods outperform most of the fusion based methods.  Fusion methods can be divided into three broad classes: early fusion, deep fusion and late fusion, each with their own pros and cons.  While early and deep fusion have greatest potential to leverage cross modality information, they suffer from sensitivity to data alignment, often involve complicated architectures~\cite{chen2017multi,ku2018joint,xu2018pointfusion,liang2018deep}, and typically require pixel-level correspondences of sensor data.  On the other hand, late fusion systems are much simpler to build as they incorporate pre-trained, single-modality detectors without change, an only need association at the detection level.  Our late fusion approach uses much-reduced thresholds for each sensor and combines detection candidates before Non-Maximum Suppression (NMS).  By leveraging cross-modality information, it can keep detection candidates that would be mistakenly suppressed by single-modality methods.

We propose Camera-LiDAR Object Candidates Fusion (CLOCs) as a way to achieve improved accuracy for 3D object detection. The proposed architecture delivers the following contributions:
\begin{itemize}
\item  \textbf{Versatility \& Modularity}: CLOCs uses any pair of pre-trained 2D and 3D detectors without requiring re-training, and hence, can be readily employed by any relevant already-optimized detection approaches. 
\item  \textbf{Probabilistic-driven Learning-based Fusion}: CLOCs is designed to exploit the geometric and semantic consistencies between 2D and 3D detections and automatically learns probabilistic dependencies from training data to perform fusion.
\item \textbf{Speed and Memory}: CLOCs is fast, leveraging sparse tensors with low memory footprint, which only adds less than 3ms latency for processing each frame of data on a desktop-level GPU.
\item \textbf{Detection Performance}: CLOCs improves single-modality detectors, including state-of-the-art detectors, to achieve new performance levels.  At time of submission, CLOCs ranks the highest among all the fusion based methods in the official KITTI leaderboard.
\end{itemize}

The rest of the paper is organized as follows. We first review related work in section 2. Then, we introduce the motivation of our work and why we choose to fuse the detection candidates in section 3. In section 4,  we illustrate our Camera-LiDAR Object Candidates (CLOCs) Fusion architecture and relevant details of our network. We report and analyse our experimental results on the KITTI dataset in section 5. In section 6, we conclude the paper.

\section{Related Work}
The three main categories 3D object detection are based on (1) 2D images, (2) 3D point clouds and (3) both images and point clouds. Although 2D image-based methods are attractive for not requiring LiDAR, there is a large gap in 3D performance between these methods and those leveraging point clouds, and so here we focus on the latter two categories.

\subsection{3D Detection Using 2D Images}

Mousavian \textit{et al}.~\cite{mousavian20173d} leverage the geometric constraints between 2D and 3D bounding boxes to recover 3D information. \cite{chabot2017deep,mottaghi2015coarse} estimate 3D object information by calculating the similarity between 3D objects and CAD models.  \cite{wang2019pseudo} and \cite{you2019pseudo} explore using stereo images to generate dense point cloud and conduct object detection using that cloud.  These image-based methods are promising, but when compared to LiDAR-based techniques, they generate much less accurate 3D bounding boxes.

\subsection{3D Detection Using Point Cloud}

Point-cloud techniques currently lead in popularity for 3D object detection. Compared to multi-modal fusion based methods, single sensor setup avoids multi-sensor calibration and synchronization issues. However, object detection performance at longer distance is still relatively poor.   Methods vary by how they encode and learn features from raw point cloud. \cite{zhou2018voxelnet} uses voxels to encode the raw point cloud, and 3D CNNs (Convolutonal Neural Networks) are applied to learn voxel features for classification and bounding box regression. SECOND \cite{yan2018second} is the upgrade version of \cite{zhou2018voxelnet}, since raw LiDAR point cloud has very sparse data structure, it uses sparse 3D CNNs which reduces the inference time significantly. PointPillars~\cite{lang2019pointpillars} uses PointNets~\cite{qi2017pointnet} in an encoder that represents point clouds organized in vertical columns (pillars) followed with a 2D CNN detection head to perform 3D object detection; it enables inference at 62 Hz; Compared with one-stage methods discussed above, PointRCNN \cite{shi2019pointrcnn}, Fast PointRCNN \cite{Chen2019fastpointrcnn} and STD \cite{std2019yang} applies a two-stage architecture that first generate 3D proposals in a bottom-up manner and then refines these proposals in a second stage. PV-RCNN \cite{shi2020pv} leverages the advantages of both 3D voxel CNN and PointNet-based set abstraction to learn more discriminative features. Besides, Part-\(A^2\) in~\cite{shi2020part} explores predicting intra-object part locations (lower left, upper right, etc.) in the first stage, and such part locations can assist accurate 3D bounding box refinement in the second stage.

\subsection{3D Detection Using Multi-modal Fusion}
We focus on camera-LiDAR fusion methods in this section since this is the most common sensor setup for self-driving cars. Frustum PointNet \cite{qi2018frustum}, Pointfusion \cite{xu2018pointfusion} and Frustum ConvNet \cite{wang2019frustum} are the representatives of 2D driven 3D detectors, which exploit mature 2D detectors to generate 2D proposals and narrow down the 3D processing domain to the corresponding cropped region in the image. 
But the 2D image-based proposal generation might fail in some cases that could only be observed from 3D space. MV3D \cite{chen2017multi} and AVOD \cite{ku2018joint} project the raw point cloud into bird's eye view (BEV) to form a multi-channel BEV image. A deep fusion based 2D CNN is used to extract features from this BEV image as well as the front camera image for 3D bounding box regression. The overall performance of these fusion based methods is worse than LiDAR-only based methods. Possible reasons include: First, transforming raw point cloud into BEV image loses spatial information. Second, the crop and resize operation used in these algorithms in order to fuse feature vectors from different sensor modalities may destroy the feature structure from each sensor. Camera images are high-resolution dense data, while LiDAR point cloud are low-resolution sparse data, fusing these two different types of data structure is not trivial. Forcing feature vectors from 2D images and 3D LiDAR point cloud to have the same size or equal-length, then concatenating, aggregating or averaging them could result in inaccurate correspondence between these feature vectors and therefore is not the optimal way for fusing features. In order to fuse features from different sensor modalities with better correspondence, MMF \cite{liang2019multi} adopts continuous convolution \cite{liang2018deep} to build dense LiDAR BEV feature maps and do point-wise feature fusion with dense image feature maps. MMF is currently one of the best public multi-modal fusion based 3D detector according to the KITTI 3D/BEV object detection benchmark. However, it is still 2$\sim$4\% worse in moderate level than the best LiDAR-only based detectors in KITTI leaderboard.

\section{Motivation}

\subsection{2D and 3D Object Detection}
\begin{figure}[htp]
    \centering
    \includegraphics[width=0.8\columnwidth]{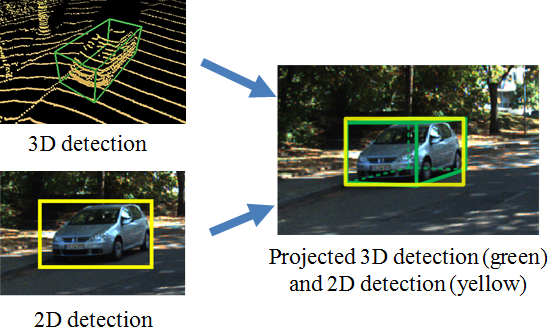}
    \caption{2D and 3D object detection.  An object that is correctly detected by both a 2D and 3D detector will have highly overlapped bounding boxes in the image plane.}
    \label{fig2}
    \vspace{-0.5cm}
\end{figure}

We first introduce the basic concepts of 2D and 3D object detection used in this paper. 2D detection systems discussed in this paper take RGB images as input, and output classified 2D axis-aligned bounding boxes with confidence scores, as shown in Fig \ref{fig2}. 3D detection systems generate classified oriented 3D bounding boxes with confidence scores, as shown in Fig \ref{fig2}.  In the KITTI dataset~\cite{geiger2012we} only rotation in z axis is considered (yaw angle), while rotations in x and y axis is set to zero for simplicity.  Using calibration parameters of the camera and LiDAR, the 3D bounding box in the LiDAR coordinate can be accurately projected into the image plane, as shown in Fig \ref{fig2}. 

\subsection{Why Fusion of Detection Candidates}
Fusion architectures can be categorized based on at what point during their processing features from different modalities are combined. Three general categories are (1) {\em early fusion} which combines data at the input, (2) {\em deep fusion} which has different networks for different modalities while simultaneously combining intermediate features, and (3) {\em late fusion} which processes each modality on a separate path and fuses the outputs in the decision level.


Early fusion has the greatest opportunity for cross-modal interaction, but at the same time inherent data differences between modalities including alignment, representation, and sparsity are not necessarily well-addressed by passing them all through the same network.  

Deep fusion addresses this issue by including separate channels for different modalities while still combining features during processing.  This is the most complicated approach, and it is not easy to determine whether or not the complexity actually leads to real improvements; simply showing gain over single-modality methods is insufficient.

Late fusion has a significant advantage in training; single modality algorithms can be trained using their own sensor data. Hence, the multi-modal data does not need to be synchronized or aligned with other modalities.  Only the final fusion step requires jointly aligned and labeled data.  Additionally, the detection candidate data that late fusion operates on is compact and simple to encode for a network. Since late fusion prunes rather than creates new detections, it is important that the input detectors be tuned to maximize their recall rate rather than their precision.  In practice, this implies that individual modalities (a) avoid the NMS stage, which may mistakenly suppress true detections. and (b) keep thresholds as low as possible.



In our late fusion framework, we incorporate all detection candidates before NMS in the fusion step to maximize the probability of extracting all potential correct detections.  Our approach is data-driven; we train a discriminative network that receives as input the output scores and classifications of individual detection candidates, as well as spatial descriptions of the detection candidates.  It learns from data how best to combine input detection candidates for a final output detection.

\section{Camera-LiDAR Object Candidates Fusion}

\subsection{Geometric and Semantic Consistencies}

For a given frame of image and LiDAR data there may be many detection candidates of with various confidences in each modality from which we seek a single set of 3D detections and scores.  Fusing these detection candidates requires an association between the different modalities (even if the association is not unique).  For this we build a geometric association score and apply semantic consistency.  These are described in more detail as follows.

\noindent
{\bf Geometric consistency} ~ An object that is correctly detected by both a 2D and 3D detector will have an identical bounding box in the image plane, see Fig \ref{fig2}, whereas false positives are less likely to have identical bounding boxes.  Small errors in pose will result in a reduction of overlap.  This motivates an image-based Intersection over Union (IoU) of the 2D bounding box and the bounding box of the projected corners of the 3D detection, to quantify geometric consistency between a 2D and a 3D detection.   

\noindent
{\bf Semantic consistency} ~ Detectors may output multiple categories of objects, but we only associate detections of the same category during fusion.  We avoid thresholding detections at this stage (or use very low thresholds), and leave thresholding to the final output based on the final fused score. 

The two types of consistencies illustrated above is the fundamental concept used in our fusion network.

\begin{figure*}[t]
    \centering
    \includegraphics[width=1.8\columnwidth]{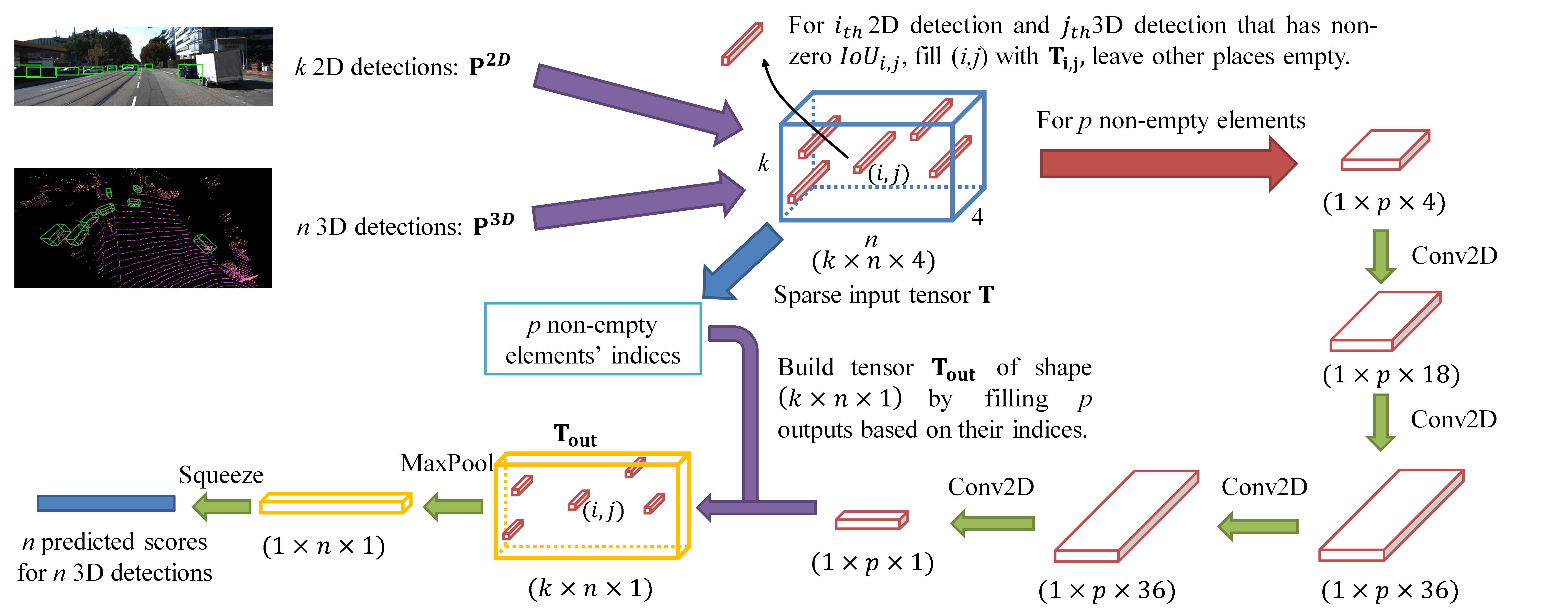}
    \caption{CLOCs Fusion network architecture. First, individual 2D and 3D detection candidates are converted into a set of consistent joint detection candidates (a sparse tensor, the blue box); Then a 2D CNN is used to process the non-empty elements in the sparse input tensor; Finally, this processed tensor is mapped to the desired learning targets, a probability score map, through maxpooling.}
    \label{figure4}
\end{figure*}

\subsection{Network Architecture}
In this section we explain the preprocessing/encoding of fused data, the fusion network architecture and the loss function used for training.

\subsubsection{Sparse Input Tensor Representation}
The goal of our encoding step is to convert all individual 2D and 3D detection candidates into a set of all consistent joint detection candidates which can be fed into our fusion network. The general output of a 2D object detector are a set of 2D bounding boxes in the image plane and corresponding confident scores. For $k$ 2D detection candidates in one image can be defined as follows:

\begin{align}
\begin{split}
    \mathbf{P^{2D}}=&\{\mathbf{p^{2D}_1},\mathbf{p^{2D}_2},...\mathbf{p^{2D}_k}  \}, \\
    \mathbf{p^{2D}_i}=&\{ \left[x_{i1},y_{i1},x_{i2},y_{i2}  \right],s^{2D}_{i} \}
\end{split}
\end{align} 

\noindent
$\mathbf{P^{2D}}$ is the set of all $k$ detection candidates in one image, for $i_{th}$ detection $\mathbf{p^{2D}_i}$, $x_{i1},y_{i1}$ and $x_{i2},y_{i2}$ are the pixel coordinates of the top left and bottom right corner points from the 2D bounding box. $s^{2D}_{i}$ is the confident score. 

The output of 3D object detectors are 3D oriented bounding boxes in LiDAR coordinate and confident scores. There are multiple ways to encode the 3D bounding boxes, in KITTI dataset \cite{geiger2012we}, a 7-digit vector containing 3D dimension (height, width and length), 3D location (x,y,z) and rotation (yaw angle) is used. For $n$ 3D detection candidates in one LiDAR scan can be defined as follows:
\begin{align}
\begin{split}
    \mathbf{P^{3D}}=&\{\mathbf{p^{3D}_1},\mathbf{p^{3D}_2},...\mathbf{p^{3D}_n}  \}, \\
    \mathbf{p^{3D}_i}=&\{ \left[h_{i},w_{i},l_{i},x_{i},y_{i},z_{i},\theta_{i}  \right], s^{3D}_{i} \}
\end{split}
\end{align} 
\noindent
where $\mathbf{P^{3D}}$ is the set of all $n$ detection candidates in one LiDAR scan, for $i_{th}$ detection $\mathbf{p^{3D}_i}$, $\left[h_{i},w_{i},l_{i},x_{i},y_{i},z_{i},\theta_{i}\right]$ is the 7-digit vector for 3D bounding box. $s^{3D}_{i}$ is the 3D confident score. Note that we take 2D and 3D detections without doing NMS, as discussed in the previous section, some correct detections may be suppressed because of limited information from single sensor modality. Our proposed fusion network would reevaluate all detection candidates from both sensor modalities to make better predictions. For $k$ 2D detections and $n$ 3D detections, we build a $k \times n \times 4$ tensor $\mathbf{T}$, as shown in Fig \ref{figure4}. For each element $\mathbf{T_{i,j}}$, there are 4 channels denoted as follows:
\begin{align}
    \mathbf{T_{i,j}}=\{ IoU_{i,j},s^{2D}_{i},s^{3D}_{j},d_{j} \}
\end{align}
\noindent
where $IoU_{i,j}$ is the $IoU$ between $i_{th}$ 2D detection and $j_{th}$ projected 3D detection, $s^{2D}_{i}$ and $s^{3D}_{j}$ are the confident scores for $i_{th}$ 2D detection and $j_{th}$ 3D detection respectively. $d_{j}$ represents the normalized distance between the $j_{th}$ 3D bounding box and the LiDAR in $xy$ plane. Elements $\mathbf{T_{i,j}}$ with zero $IoU$ are eliminated as they are geometrically inconsistent.

The input tensor $\mathbf{T}$ is sparse because for each projected 3D detection, only few 2D detections intersect with it and so most elements are empty. The fusion network only needs to learn from these intersected examples. Because we take the raw predictions before NMS, $k$ and $n$ are large numbers, for SECOND \cite{yan2018second}, there are 70400 ($200 \times 176 \times 2$) predictions in each frame. It would be impractical to do $1 \times 1$ convolution on a dense tensor with this shape. We propose an implementation architecture to utilize the sparsity of tensor $\mathbf{T}$ and make the calculations much faster and feasible for large $k$ and $n$ values. Only non-empty elements are delivered to the fusion network for processing, shown in Fig. \ref{figure4}. As we would discuss later, the indices of the non-empty elements ($i,j$) are important for further calculations, therefore the indices of these non-empty elements are saved in the cache, as shown in the blue box in Fig. \ref{figure4}. Here noted that for projected 3D detection $\mathbf{p_j}$ that has no 2D detection intersected, we still fill the last element in $j_{th}$ column $\mathbf{T_{k,j}}$ in $\mathbf{T}$ with the available 3D detection information and set $IoU_{k,j}$ and $s^{2D}_{k}$ as -1. Because sometimes 3D detector could detect some objects that 2D detector couldn't and we do not want to discard these 3D detections. Setting the $IoU$ and $s^{2D}$ to -1 rather than 0 enables our network to distinguish this case from other examples with very small $IoU$ and $s^{2D}$.


\definecolor{LightCyan}{rgb}{0.88,1,1}
\begin{table*}[t]
\setlength{\tabcolsep}{5pt}
\caption{Performance comparison of object detection with state-of-the-art camera-LiDAR fusion methods on car class of KITTI test set (new 40 recall positions metric).  CLOCs fusion improves the baselines and outperforms other state-of-the-art fusion-based detectors. 3D and bird's eye view detection are evaluated by Average Precision (AP)}
\label{table_1}
\begin{center}
\begin{tabular}{|c|c|c|c|c|c|c|c|}
\hline
\multirow{2}{*}{Detector} & \multirow{2}{*}{Input Data} &  \multicolumn{3}{|c|}{3D AP (\%)} & \multicolumn{3}{|c|}{Bird's Eye View AP (\%)}\\\cline{3-8}
& & easy & moderate & hard & easy & moderate & hard\\
\hline
SECOND (baseline) \cite{yan2018second} & LiDAR & 83.34 & 72.55 & 65.82 & 89.39 & 83.77 & 78.59  \\
\hline
\makecell{CLOCs\_SecCas \\(SECOND+Cascad R-CNN)} & LiDAR+Img & 86.38 & 78.45 & 72.45 & 91.16 & 88.23 & 82.63 \\
\hline
\rowcolor{LightCyan}
\makecell{\textit{Improvement} \\ \textit{(CLOCs\_SecCas over SECOND)}} & - & \textit{+3.04} & \textit{+5.90} & \textit{+6.63} & \textit{+1.77} & \textit{+4.46} & \textit{+4.04} \\
\hline
PointRCNN (baseline) \cite{shi2019pointrcnn} & LiDAR   & 86.23 & 75.81 & 68.99 & 92.51 & 86.52 & 81.39  \\
\hline
\makecell{CLOCs\_PointCas \\(PointRCNN+Cascad R-CNN)} & LiDAR+Img & 87.50 & 76.68 & 71.20 & 92.60 & 88.99 & 81.74 \\
\hline
\rowcolor{LightCyan}
\makecell{\textit{Improvement} \\ \textit{(CLOCs\_PointCas over PointRCNN)}} & - & \textit{+1.27} & \textit{+1.04} & \textit{+2.21} & \textit{+0.09} & \textit{+2.47} & \textit{+0.35} \\
\hline
PV-RCNN (baseline) \cite{shi2020pv} & LiDAR   & 87.45 & 80.28 & 76.21 & 91.91 & 88.13 & 85.41  \\
\hline
\makecell{CLOCs\_PVCas \\(PV-RCNN+Cascad R-CNN)} & LiDAR+Img   & \textbf{88.94} & \textbf{80.67} & \textbf{77.15} & 93.05 & \textbf{89.80} & \textbf{86.57}  \\
\hline
\rowcolor{LightCyan}
\makecell{\textit{Improvement} \\ \textit{(CLOCs\_PVCas over PV-RCNN)}} & - & \textit{+1.49} & \textit{+0.39} & \textit{+0.94} & \textit{+1.14} & \textit{+1.67} & \textit{+1.17} \\
\hline
F-PointNet \cite{qi2018frustum} & LiDAR+Img & 82.19 & 69.79 & 60.59 &91.17 & 84.67 & 74.77 \\
\hline
AVOD-FPN \cite{ku2018joint} & LiDAR+Img & 83.07 & 71.76 & 65.73 & 90.99 & 84.82 & 79.62  \\
\hline
F-ConvNet \cite{wang2019frustum} & LiDAR+Img  & 87.36 & 76.39 & 66.69 & 91.51 & 85.84 & 76.11  \\
\hline
UberATG-MMF \cite{liang2019multi} & LiDAR+Img & 88.40 & 77.43 & 70.22 & 93.67 & 88.21 & 81.99  \\
\hline
UberATG-ContFuse \cite{liang2018deep} & LiDAR+Img  & 83.68 & 68.78 & 61.67 & \textbf{94.07} & 85.35 & 75.88  \\
\hline
\end{tabular}
\end{center}
\vspace{-0.5cm}
\end{table*}

\subsection{Network Details}
The fusion network is a set of $1\times1$ 2D convolution layers. We use Conv2D($c_{in},c_{out},\mathbf{k},\mathbf{s}$) to represent an 2 dimensional convolution operator where $c_{in}$ and $c_{out}$ are the number of input and output channels, $\mathbf{k}$ and $\mathbf{s}$ are the kernel size vector and stride respectively. We employ four convolution layers sequentially as Conv2D(4, 18, (1,1), 1), Conv2D(18, 36, (1,1), 1), Conv2D(36, 36, (1,1), 1) and Conv2D(36, 1, (1,1), 1), which yields a tensor of size $1 \times p \times 1$ shown in Fig.~\ref{figure4}, where $p$ is the number of non-empty elements in the input tensor $\mathbf{T}$. Note that for the first three convolution layers, after each convolution layer applied, ReLU \cite{nair2010rectified} is used. Since we have saved the indices of these non-empty elements ($i,j$), as shown in Fig.~\ref{figure4} now we could build a tensor $\mathbf{T_{out}}$ of shape $k \times n \times 1$ by filling $p$ outputs based on the indices $(i,j)$ and putting negative infinity elsewhere. Finally, this tensor is mapped to the desired learning targets, a probability score map of size $1 \times n$, through maxpooling in the first dimension. 

\subsection{Loss}
We use a cross entropy entropy loss for target classification, modified by the focal loss in~\cite{lin2017focal}
with parameters $\alpha=0.25$ and $\gamma=2$ to address the large class imbalance between targets and background.

\subsection{Training}
The fusion network is trained using stochastic gradient descent (SGD). We use the Adam optimizer with an initial learning rate of 3 * $10^{-3}$ and decay the learning rate by a factor of 0.8 for 15 epochs.

\section{Experimental Results}
In this section we present our experimental setup and results, including dataset, platform, performance results and analyses. For all experiments, we focus on the car class since it has the most training and testing samples in the KITTI \cite{geiger2012we} dataset.
\subsection{Dataset}
Our fusion system is evaluated on the challenging 3D object detection benchmark KITTI dataset \cite{geiger2012we} which has both LiDAR point clouds and camera images. There are 7481 training samples and 7518 testing samples. Ground truth labels are only available for training samples. For the evaluation of testing samples, one needs to submit the detection results to KITTI server. For experimental studies, we follow the convention in \cite{chen20153d} to split the original training samples into 3712 training samples and 3769 validation samples. We compare our method with sate-of-the-art multi-modal fusion methods of 3D object detection on official test split of KITTI as well as validation split.

\subsection{2D/3D Detector Setup}
We apply our fusion network for a combination of different 2D and 3D detectors to demonstrate the flexibility of our proposed pipeline. The 2D detectors we used are: RRC \cite{ren2017accurate}, MS-CNN \cite{cai2016unified} and Cascade R-CNN \cite{cai2019cascade}. The 3D detectors we incorporated are: SECOND \cite{yan2018second}, PointPillars \cite{lang2019pointpillars}, PointRCNN \cite{shi2019pointrcnn} and PV-RCNN \cite{shi2020pv}.  While not the top performers within the KITTI leaderboard, we have selected these methods as they are the best currently-available open-source detectors. Our experiments show that CLOCs improves the performance of these detectors significantly. At the time of submission, CLOCs fusion of PV-RCNN with Cascade R-CNN, is ranked number 4 on KITTI 3D detection leaderboard, number 6 on Bird Eye View detection leaderboard, number 1 on 2D detection leaderboard, and outperforms all other fusion methods. 

\begin{table}[htbp]
    \setlength{\tabcolsep}{1.5pt}
    \caption{3D and bird's eye view performance of fusion with different combinations of 2D/3D detectors through CLOCs on car class of KITTI validation set (new 40 recall positions metric). Our CLOCs fusion methods outperform the baseline methods.}
    \label{table_2}
    \begin{center}
        \begin{tabular}{|c|c|c|c|c|c|c|}
        \hline
        \multirow{2}{*}{Detector} & \multicolumn{3}{|c|}{3D AP (\%)} & \multicolumn{3}{|c|}{Bird's Eye View AP (\%)} \\\cline{2-7}
        & easy & moderate & hard & easy & moderate & hard\\
        \hline
        SECOND (baseline) & 90.97 & 79.94 & 77.09 & 95.61 & 89.54 & 86.96 \\ 
        \hline
        SECOND+RRC & 92.69 & 82.69 & 78.20 & 96.53 & 92.78 & 87.74 \\ 
        SECOND+MSCNN & 92.37 & 82.36 & 78.23 & 96.34 & 92.59 & 87.81 \\ 
        SECOND+C-RCNN* & 92.35 & 82.73 & 78.10 & 96.34 & 92.57 & 89.36 \\ 
        \hline
        \hline
        PointPillars (baseline) & 87.37 & 76.17 & 72.88 & 92.40 & 87.79 & 86.39 \\ 
        \hline
        PointPillars+RRC & 88.48 & 78.50 & 75.13 & 93.53 & 88.87 & 87.09 \\ 
        PointPillars+MSCNN & 89.22 & 77.05 & 73.16 & 92.80 & 88.46 & 87.26 \\ 
        PointPillars+C-RCNN* & 89.95 & 78.99 & 73.27 & 93.77 & 88.27 & 87.34 \\ 
        \hline
        \hline
        PointRCNN (baseline) & 92.54 & 82.16 & 77.88 & 95.58 & 88.78 & 86.34 \\ 
        \hline
        PointRCNN+RRC & 92.67 & 84.75 & 81.82 & 95.98 & 90.80 & 87.96 \\ 
        PointRCNN+MSCNN & 92.64 & 83.26 & 79.88 & 95.60 & 90.05 & 87.05 \\ 
        PointRCNN+C-RCNN* & 93.09 & 84.09 & 80.73 & 96.13 & 90.19 & 87.26 \\ 
        \hline
        \hline
        PV-RCNN (baseline) & 92.10 & 84.36 & 82.48 & 93.02 & 90.33 & 88.53 \\ 
        \hline
        PV-RCNN+RRC & 92.82 & 85.59 & 83.00 & 93.65 & 92.40 & 90.19 \\ 
        PV-RCNN+MSCNN & 92.66 & 83.89 & 83.29 & 93.50 & 91.63 & 89.42 \\ 
        PV-RCNN+C-RCNN* & 92.78 & 85.94 & 83.25 & 93.48 & 91.98 & 89.48 \\ 
         \hline
        \end{tabular}
        \begin{tablenotes}
            \small
            \item *C-RCNN is Cascade R-CNN.
        \end{tablenotes}
    \end{center}
    \vspace{-0.5cm}
\end{table}

\begin{table}[b]
    \setlength{\tabcolsep}{2pt}
    \caption{3D and bird's eye view performance of fusion on pedestrian class of KITTI validation set (using new 40 recall positions). Our CLOCs fusion methods outperform the corresponding baseline methods}
    \label{table_3}
    \begin{center}
        \begin{tabular}{|c|c|c|c|c|c|c|}
        \hline
        \multirow{2}{*}{Detector} & \multicolumn{3}{|c|}{3D AP (\%)} & \multicolumn{3}{|c|}{Bird's Eye View AP (\%)} \\\cline{2-7}
        & easy & moderate & hard & easy & moderate & hard\\
        \hline
        SECOND (baseline) & 58.01 & 51.88 & 47.05 & 61.97 & 56.77 & 51.27 \\ 
        \hline
        SECOND+MSCNN & 62.54 & 56.76 & 52.26 & 69.35 & 63.47 & 58.93 \\
        \hline
        \hline
        PointPillars (baseline) & 58.38 & 51.42 & 45.20 & 66.97 & 59.45 & 53.42  \\ 
        \hline
        PointPillars+MSCNN & 60.33 & 54.17 & 46.42 & 69.29 & 63.00 & 54.80 \\ 
        \hline
        \end{tabular}
    \end{center}
\end{table}

\begin{table}[htbp]
    \setlength{\tabcolsep}{2pt}
    \caption{Performance of fusion on cyclist class of KITTI validation set (new 40 recall positions). Our CLOCs fusion methods outperform the corresponding baseline methods}
    \label{table_4}
    \begin{center}
        \begin{tabular}{|c|c|c|c|c|c|c|}
        \hline
        \multirow{2}{*}{Detector} & \multicolumn{3}{|c|}{3D AP (\%)} & \multicolumn{3}{|c|}{Bird's Eye View AP (\%)} \\\cline{2-7}
        & easy & moderate & hard & easy & moderate & hard\\
        \hline
        SECOND (baseline) & 78.50 & 56.74 & 52.83 & 81.91 & 59.36 & 55.53 \\ 
        \hline
        SECOND+MSCNN & 85.47 & 59.47 & 55.00 & 88.96 & 63.40 & 59.81 \\
        \hline
        \hline
        PointPillars (baseline) & 82.31 & 59.33 & 55.25 & 84.65 & 61.39 & 57.28 \\ 
        \hline
        PointPillars+MSCNN & 90.26 & 64.84 & 59.59 & 92.64 & 67.97 & 62.31 \\ 
        \hline
        \end{tabular}
    \end{center}
\end{table}

\begin{figure}[htbp]
    \begin{subfigure}{0.49\columnwidth}
        \includegraphics[width=\columnwidth,trim={0.2cm 0.1cm 0.2cm 0.1cm}]{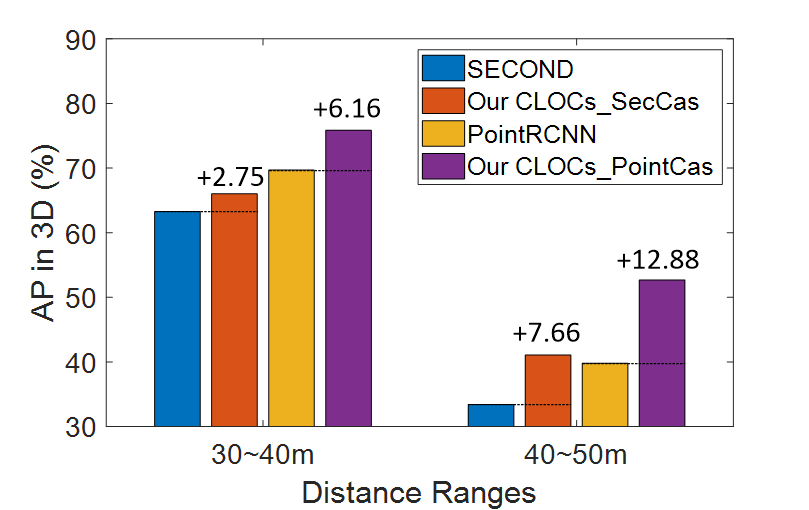}
        \caption{3D detection}
        \label{distance_a}
    \end{subfigure}
    \begin{subfigure}{0.49\columnwidth}
        \includegraphics[width=\columnwidth,trim={0.2cm 0.1cm 0.25cm 0.1cm}]{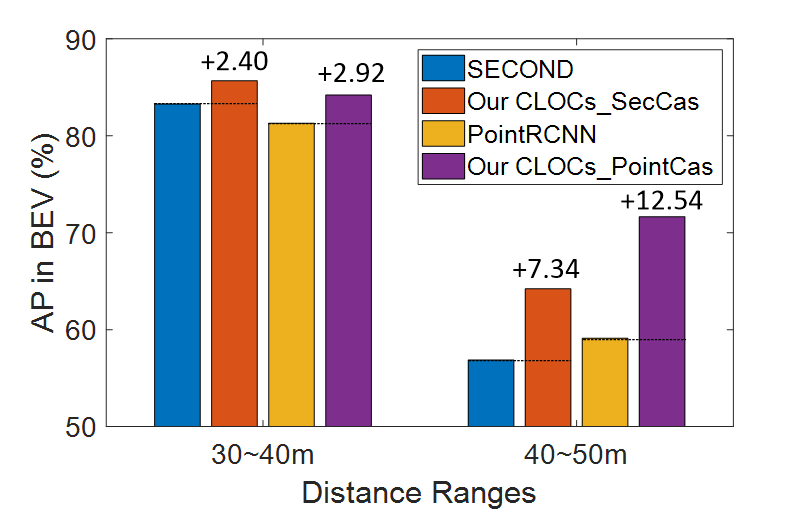}
        \caption{Bird's eye view detection}
        \label{distance_b}
    \end{subfigure}
\caption{Average Precision (AP) based on distance. CLOCs\_SecCas, CLOCs\_PointCas and their corresponding baselines SECOND and PointRCNN are shown in this figure. Our CLOCs outperforms the baseline by a large margin especially in long distance ($40\sim50m$).}
\label{distance_analyses}
\vspace{-0.5cm}
\end{figure}

\begin{figure*}[ht]
    \includegraphics[width=2\columnwidth]{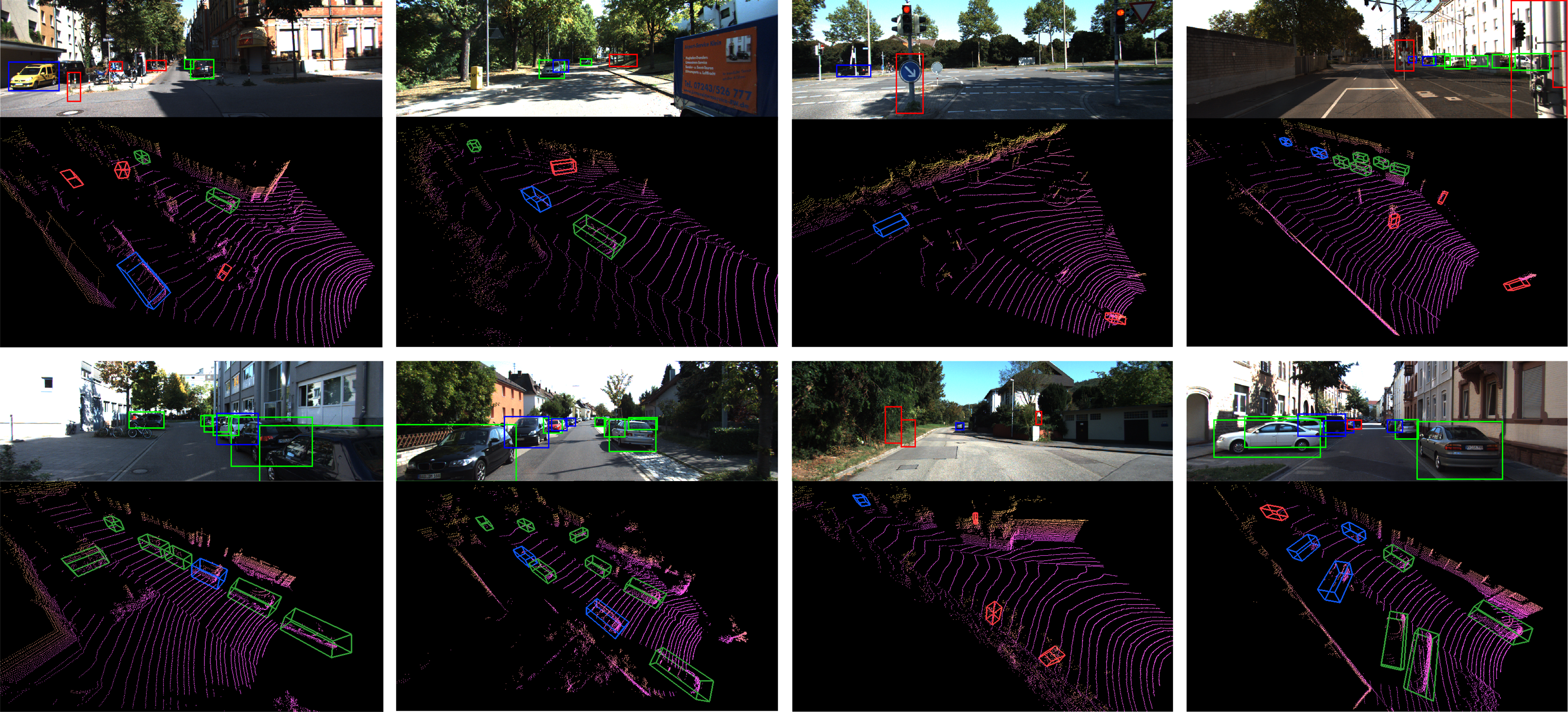}
\caption{Qualitative results of our CLOCs on KITTI test set compared to SECOND \cite{yan2018second}. Red and blue bounding boxes are false and missed detections from SECOND respectively that are corrected by our CLOCs. Green bounding boxes are correct detections. The upper row in each image is the 3D detection projected to image, the others are 3D detections in LiDAR point clouds.}
\label{Qualitative_results}
\vspace{-0.5cm}
\end{figure*}

\subsection{Evaluation Results}
We evaluate the detection results on the KITTI test server. The IoU threshold for car is 0.7. All the instances are classified into three difficulty levels: easy, moderate and hard, based on their 2D bounding boxes' height, occlusion level and truncation level. Since KITTI has some restrictions on the number of submissions, we only show the results evaluated on the official KITTI test server from three fusion combinations of 2D and 3D detectors, which are SECOND \cite{yan2018second} and Cascade R-CNN \cite{cai2019cascade}, written as CLOCs\_SecCas, PointRCNN \cite{shi2019pointrcnn} and Cascade R-CNN, as CLOCs\_PointCas, PV-RCNN \cite{shi2020pv} and Cascade R-CNN, as CLOCs\_
PVCas. All the other combinations are evaluated on the validation set. 
Table \ref{table_1} shows the performance of our fusion method on the KITTI test set through server submission.  Our methods outperform all multi-modal fusion based works in moderate and hard level at the time of submission. Note that the official open-source code of PV-RCNN performs slightly worse than the private one owned by the PV-RCNN authors shown on the KITTI leaderboard, and our CLOCs\_PVCas result is based on the open-source PV-RCNN. The baseline PV-RCNN in Table~\ref{table_1} refers to the open-source PV-RCNN. As shown in Table~\ref{table_1}, compared to baseline methods SECOND, PointRCNN and PV-RCNN, fusion with Cascade R-CNN through our fusion network increases the performance in 3D and BEV object detection by a large margin.

We evaluate the performance of all the combinations of 2D and 3D detectors on car class of KITTI validation set, the results are shown in Table \ref{table_2}. Compared to the corresponding baseline 3D detectors, our fusion methods have better performance in 3D and BEV detection benchmark. These results demonstrate the effectiveness as well as the flexibility of our fusion approach.

Table \ref{table_3} and Table \ref{table_4} show the 3D and BEV evaluation results of pedestrian and cyclist on KITTI validation set. The IoU threshold for pedestrian and cyclist is 0.5. Here for 3D detectors, only SECOND \cite{yan2018second} and PointPillars \cite{lang2019pointpillars} publish their training configurations for class pedestrian and cyclist; for 2D detectors, only MSCNN \cite{cai2016unified} does. Therefore, we only show the evaluation results based on SECOND, PointPillars and MSCNN. As shown in Table \ref{table_3} and Table \ref{table_4}, our fusion method improves the detection performance by a large margin.

Fig. \ref{distance_analyses} shows the average precision (AP) on KITTI validation set in different distance ranges. The distance is defined as the Euclidean distance in $xy$ plane between objects and LiDAR. The blue bars are the APs for SECOND detector, the orange bars represent APs for our CLOCs\_SecCas. The yellow and purple bars show the APs of PointRCNN and CLOCs\_PointCas respectively. As shown in Fig. \ref{distance_analyses}, APs for CLOCs is higher than the corresponding baselines in all distance ranges on both 3D and BEV detection benchmarks. The largest improvement is in $40\sim50m$. This is because the point clouds in long distance are too sparse for LiDAR-only detectors such as SECOND and PointRCNN, while CLOCs could utilize 2D detections to improve the performance.

Fig. \ref{Qualitative_results} shows some qualitative results of our proposed fusion method on the KITTI \cite{geiger2012we} test set. Red bounding boxes represent wrong detections (false positives) from SECOND that are deleted by our CLOCs, blue bounding boxes stand for missed detections from SECOND that are corrected by our CLOCs, green bounding boxes are correct detections.

\begin{table}[t]
    \setlength{\tabcolsep}{1.5pt}
    \caption{Performance of CLOCs with PointRCNN using different score scales on car class of KITTI validation set. Because the sigmoid score from PointRCNN poorly approximates probability of a target (or precision), using it for fusion could result in worse performance.}
    \label{table_5}
    \begin{center}
        \begin{tabular}{|c|c|c|c|c|c|c|}
        \hline
        \multirow{2}{*}{Type of Scores} & \multicolumn{3}{|c|}{3D AP (\%)} & \multicolumn{3}{|c|}{Bird's Eye View AP (\%)} \\\cline{2-7}
        & easy & moderate & hard & easy & moderate & hard\\
        \hline
        log score & 93.09 & 84.09 & 80.73 & 96.13 & 90.19 & 87.26 \\ 
        \hline
        sigmoid score & 91.64 & 82.96 & 79.13 & 95.33 & 89.70 & 86.36 \\
        \hline
        \hline
        corrected sigmoid score & 92.83 & 83.73 & 80.12 & 95.88 & 90.19 & 87.08  \\ 
        \hline
        corrected log score & 92.88 & 83.92 & 80.22 & 96.07 & 89.93 & 87.21 \\ 
        \hline
        \end{tabular}
    \end{center}
    \vspace{-0.8cm}
\end{table}

\subsection{Score Scales}

There are two common output scores for detectors: the first is a real number approximating the log likelihood ratio between target and clutter, and the second is a sigmoid transformation of this onto the range 0 to 1, so approximating a probability of target.  We compare use of these in CLOCs in Table \ref{table_5} and find improved performance using the log likelihood score.  The primary reason for the poor performance for the normalized score is that it poorly approximates a probability of target (or precision).  Using this score forces the fusion network to learn a non-linear correction, whereas the equivalent log likelihood score discrepancy is a simple offset that can easily corrected by the fusion layer.  If we instead use a fitted sigmoid to obtain better probabilistic outputs from the PointRCNN, then fusion works equally well with either input.  In general we believe it is simpler to use a log likelihood output for each single-modality detector and fuse these.

\subsection{Ablation Study}

We evaluate the contribution of each channel and focal loss in our fusion pipeline. The four channel includes: IoU between 2D detections and projected 3D detections ($IoU$), 2D confident score ($s^{2D}$), 3D confident score ($s^{3D}$) and normalized distance ($d$) between 3D bounding box and the LiDAR in $xy$ plane. The results are shown in Table \ref{table_6}.  

$IoU$, as the measure of geometric consistency, is crucial to the fusion network.  Without $IoU$, the association between 2D and 3D detections would be ambiguous and further lead to degrade performance.  2D confident score indicates the certainty of 2D detections, which could provide useful clues for the fusion.  3D confident score ($s^{3D}$) plays the most important role among the four channels, because CLOCs generates new confident scores to all 3D detection candidates through fusion in which original 3D scores are highly important evidences.  Closer objects usually are easier detected because there are more hits from LiDAR, the normalized distance ($d$) could be a useful indicator for this.  Because there is a large imbalance between positives and negatives among the detection candidates, focal loss could address this issue and improve the detection accuracy.

\begin{table}[t]
    \setlength{\tabcolsep}{1pt}
    \caption{The contribution of each channel and focal loss in our CLOCs fusion pipeline. The results are on the moderate level car class of KITTI $val$ split with AP calculated by 40 recall positions. SECOND and Cascade R-CNN are fused in this experiment, so the baseline model is SECOND}
    \label{table_6}
    \begin{center}
        \begin{tabular}{|>{\centering\arraybackslash}X m{0.75cm}|>{\centering\arraybackslash}X  m{0.75cm}|>{\centering\arraybackslash}X m{0.75cm}|>{\centering\arraybackslash}X m{0.75cm}|>{\centering\arraybackslash}X m{1.25cm}|c|c|}
        \hline
        \hline
        $IoU$ & $s^{2D}$ & $s^{3D}$ & $d$ & focal loss & 3D AP & BEV AP  \\ 
        \hline
        \hline
          &  &  &  &  & 79.94  &  89.54 \\
        \hline
          & \checkmark & \checkmark & \checkmark & \checkmark & 78.95 & 88.43  \\
        \hline
        \checkmark &  & \checkmark & \checkmark & \checkmark & 80.96 & 90.32  \\ 
        \hline
        \checkmark & \checkmark &  & \checkmark & \checkmark & 38.64 & 47.16  \\
        \hline
        \checkmark & \checkmark & \checkmark &  & \checkmark & 81.96 & 91.90  \\
        \hline
        \checkmark & \checkmark & \checkmark & \checkmark &  & 81.01 & 92.17  \\
        \hline
        \checkmark & \checkmark & \checkmark & \checkmark & \checkmark & \textbf{82.73} & \textbf{92.57}  \\
        \hline
        \end{tabular}
    \end{center}
    \vspace{-0.5cm}
\end{table}

\section{CONCLUSIONS}

In this paper, we propose Camera-LiDAR Object Candidates  Fusion (CLOCs), as a fast and simple way to improve performance of just about any 2D and 3D object detectors when both LiDAR and camera data are available.  CLOCs exploits the geometric and semantic consistencies between 2D and 3D detections and automatically learns fusion parameters.   The experiments show that our fusion method outperforms previous state-of-the-art methods by a large margin on the challenging 3D detection benchmark of KITTI dataset, especially in long distance detection.  As such, CLOCs provides a baseline for other types of fusion including early and deep fusion.






\bibliography{IEEEexample.bib}
\bibliographystyle{IEEEtran.bst}
\end{document}